%% file: main.tex
\documentclass[conference]{IEEEtran}
\usepackage{times}

\usepackage{multicol}
\usepackage[bookmarks=true]{hyperref}
\usepackage[table,xcdraw]{xcolor}
\usepackage{color}
\let\labelindent\relax
\usepackage{enumitem}
\usepackage{graphicx}
\usepackage{amsmath}
\usepackage{amsfonts}       %
\usepackage{xfrac}
\usepackage[nice]{nicefrac}
\usepackage{
tikz,
relsize,
booktabs
}
\usepackage[font={small}]{caption}
\usepackage{bbm}

\usepackage{algorithm}
\usepackage{algpseudocode}

\definecolor{citecolor}{HTML}{0071bc}
\usepackage{hyperref}
\hypersetup{breaklinks=true,colorlinks, citecolor=citecolor,linkcolor=black}

\usepackage{booktabs}       %
\usepackage{graphicx}
\usepackage{multirow}
\usepackage{longtable}
\usepackage{microtype}      %
\usepackage{tikz}

\renewcommand\thefigure{\arabic{figure}}
\newcommand{\xxnote}[3]{}
\renewcommand{\xxnote}[3]{\color{#2}{#1: #3}}

\newcommand{\method}{\texttt{OK-Robot}}

\newcommand{\numhomes}{10}

\newcommand{\numhomeexperiments}{171}

\newcommand{\homesuccessrate}{58.5\%}
\newcommand{\cleanhomesuccessrate}{82.4\%}

\newcommand{\mysection}[1]{%
  \vspace{0.25\baselineskip}
  \par\noindent\textbf{#1}:%
}

\newcommand{\website}{https://ok-robot.github.io}

\IEEEoverridecommandlockouts

\usepackage[style=numeric, sorting=none, backend=bibtex, citestyle=numeric-comp, maxnames=10, minnames=3]{biblatex}%
\begin{document}
\title{\texttt{OK-Robot}:\\ What Really Matters in Integrating Open-Knowledge Models for Robotics}

\author{
Peiqi Liu*$^1$\thanks{* Denotes equal contribution and $\dagger$ denotes equal advising.}\thanks{Correspondence to: \href{mailto:mahi@cs.nyu.edu}{mahi@cs.nyu.edu}} \quad Yaswanth Orru*$^1$\quad Jay Vakil$^2$ \quad Chris Paxton$^2$ \\ Nur Muhammad Mahi Shafiullah\authorrefmark{2}$^1$ \quad Lerrel Pinto\authorrefmark{2}$^1$ \vspace{0.1in}
\\ \vspace{0.1in}
New York University$^1$, AI at Meta$^2$
\\ 
{ \tt \url{\website}}
}

\makeatletter
\let\@oldmaketitle\@maketitle%
\renewcommand{\@maketitle}{\@oldmaketitle%
    \centering
    \includegraphics[width=1.0\linewidth]{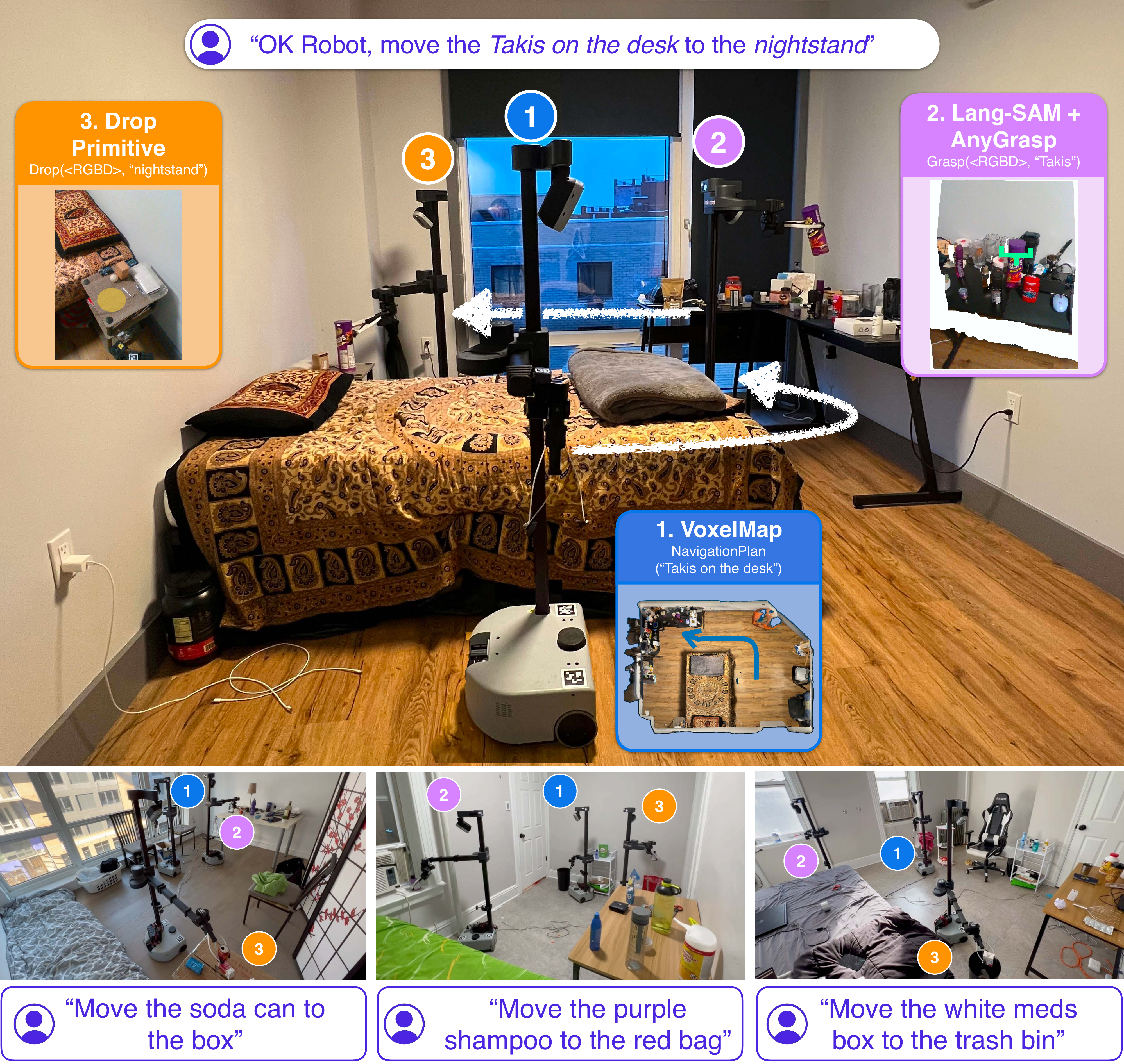}
    \captionof{figure}{\method{} is an Open Knowledge robotic system, which integrates a variety of learned models trained on publicly available data, to pick and drop objects in real-world environments. Using Open Knowledge models such as CLIP, Lang-SAM, AnyGrasp, and OWL-ViT, \method{} achieves a \homesuccessrate{} success rate across \numhomes{} unseen, cluttered home environments, and 82.4\% on cleaner, decluttered environments.}
    \label{fig:intro}
}
\makeatother

\maketitle

\input{abstract}

\IEEEpeerreviewmaketitle

\input{introduction}

\input{method}

\input{experiments}

\input{related_works}
\input{open_problems}

\section*{Acknowledgments}
NYU authors are supported by grants from Amazon, Honda, and ONR award numbers N00014-21-1-2404 and N00014-21-1-2758. NMS is supported by the Apple Scholar in AI/ML Fellowship. LP is supported by the Packard Fellowship.
Our utmost gratitude goes to our friends and colleagues who helped us by hosting our experiments in their homes.
Finally, we thank Siddhant Haldar, Paula Pascual and Ulyana Piterbarg for valuable feedback and conversations.

\printbibliography

\clearpage
\appendices
\input{appendix}
\end{document}

%% file: abstract.tex
\begin{abstract}
\label{abstract}
Remarkable progress has been made in recent years in the fields of vision, language, and robotics. We now have vision models capable of recognizing objects based on language queries, navigation systems that can effectively control mobile systems, and grasping models that can handle a wide range of objects. Despite these advancements, general-purpose applications of robotics still lag behind, even though they rely on these fundamental capabilities of recognition, navigation, and grasping. 
In this paper, we adopt a systems-first approach to develop a new Open Knowledge-based robotics framework called \method{}. By combining Vision-Language Models (VLMs) for object detection, navigation primitives for movement, and grasping primitives for object manipulation, \method{} offers a integrated solution for pick-and-drop operations without requiring any training. To evaluate its performance, we run \method{} in 10 real-world home environments. The results demonstrate that \method{} achieves a 
\homesuccessrate{} success rate in open-ended pick-and-drop tasks, representing a new state-of-the-art in Open Vocabulary Mobile Manipulation (OVMM) with nearly $1.8\times$ the performance of prior work.
On cleaner, uncluttered environments, \method{}'s performance increases to 82\%.
However, the most important insight gained from \method{} is the critical role of nuanced details when combining Open Knowledge systems like VLMs with robotic modules. 
We published our code and robot videos on \href{https://ok-robot.github.io}{https://ok-robot.github.io} to encourage further investigation.
\end{abstract}

%% file: introduction.tex
\section{Introduction}
\label{sec:intro}

Creating a general-purpose robot has been a longstanding dream of the robotics community.
With the increase in data-driven approaches and large robot models, impressive progress is being made~\cite{pinto2016supersizing,levine2018learning,ahn2022saycan, shafiullah2023bringing}. However, current systems are brittle, closed, and fail when encountering unseen scenarios.
Even the largest robotics models can often only be deployed in previously seen environments~\cite{brohan2022rt1,brohan2023rt2}. The brittleness of these systems is further exacerbated in settings where little robotic data is available, such as in unstructured home environments. %

The poor generalization of robotic systems lies in stark contrast to large vision models~\cite{zhou2022detecting,minderer2022simple,radford2021clip,marino2019ok}, which show capabilities of semantic understanding~\cite{alayrac2022flamingo, liu2023grounding, liu2023visual}, detection~\cite{zhou2022detecting,minderer2022simple}, and connecting visual representations to language~\cite{radford2019language,radford2021clip,marino2019ok}
At the same time, base robotic skills for navigation~\cite{gervet2023navigating}, grasping~\cite{sundermeyer2021contact,mahler2017dex,fang2020graspnet,fang2023anygrasp}, and rearrangement~\cite{goyal2022ifor,liu2022structdiffusion} are fairly mature. 
Hence, it is perplexing that robotic systems that combine modern vision models with robot-specific primitives perform so poorly.
To highlight the difficulty of this problem, the recent NeurIPS 2023 challenge for open-vocabulary mobile manipulation (OVMM)~\cite{homerobotovmmchallenge2023} registered a success rate of 33\% for the winning solution~\cite{melnik2023uniteam}.

So what makes open-vocabulary robotics so hard? Unfortunately, there isn't a single challenge that makes this problem hard. Instead, inaccuracies in different components compound and together results in an overall drop. For example, the quality of open-vocabulary retrievals of objects in homes is dependent on the quality of query strings, navigation targets determined by VLMs may not be reachable to the robot, and the choice of different grasping models may lead to large differences in grasping performance.
Hence, making progress on this problem requires a careful and nuanced framework that both integrates VLMs and robotics primitives, while being flexible enough to incorporate newer models as they are developed by the VLM and robotics community. 

We present \method{}, an \textit{Open Knowledge Robot} that integrates state-of-the-art VLMs with powerful robotics primitives for navigation and grasping to enable pick-and-drop. Here, \textit{Open Knowledge} refers to learned models trained on large, publicly available datasets. When placed in a new home environment, \method{} is seeded with a scan taken from an iPhone. Given this scan, dense vision-language representations are computed using LangSam~\cite{langsam2023} and CLIP~\cite{radford2021clip} and stored in a semantic memory. Then, when a language-query for an object to be picked comes in, semantic memory is queried with the language embedding to find that object. After this, navigation and picking primitives are applied sequentially to move to the desired object and pick it up. A similar process can be carried out for dropping the object.

To study \method{}, we tested it in \numhomes{} real world home environments. Through our experiments, we found that on a unseen natural home environment, a zero-shot deployment of our system achieves \homesuccessrate{} success on average. However, this success rate is largely dependant on the ``naturalness" of the environment, as we show that with improving the queries, decluttering the space, and excluding objects that are clearly adversarial (too large, too translucent, too slippery), this success rate reaches \cleanhomesuccessrate{}. Overall, through our experiments, we make the following observations:

\begin{itemize}[leftmargin=12pt]
    \item \textbf{Pre-trained VLMs are highly effective for open-vocabulary navigation:} Current open-vocabulary vision-language models such as CLIP~\cite{radford2021clip} or OWL-ViT~\cite{minderer2022simple} offer strong performance in identifing arbitrary objects in the real world, and enable navigating to them in a zero-shot manner (see Section~\ref{sec:objectnav}.)

    \item \textbf{Pre-trained grasping models can be directly applied to mobile manipulation:} Similar to VLMs, special purpose robot models pre-trained on large amounts of data can be applied out of the box to approach open-vocabulary grasping in homes. These robot models do not require any additional training or fine-tuning (see Section~\ref{sec:grasping}.)

    \item \textbf{How components are combined is crucial:} Given the pretrained models, we find that they can be combined with no training using a simple state-machine model. We also find that using heuristics to counteract the robot's physical limitations can lead to a better success rate in the real world (see Section~\ref{sec:deployment}.)

    \item \textbf{Several challenges still remain:} While, given the immense challenge of operating zero-shot in arbitrary homes, \method{} improves upon prior work, by analyzing the failure modes  we find that there are significant improvements that can be made on the VLMs, robot models, and robot morphology, that will directly increase performance of open-knowledge manipulation agents (see Section~\ref{sec:failures}).
\end{itemize}

To encourage and support future work in open-knowledge robotics, we have shared the code and modules for \method{}, and are committed to supporting reproduction of our results. More information along with robot videos and the code are available on our project website: \url{\website}.

%% file: method.tex
\setcounter{figure}{1} 
\begin{figure*}
    \centering
    \includegraphics[width=\linewidth]{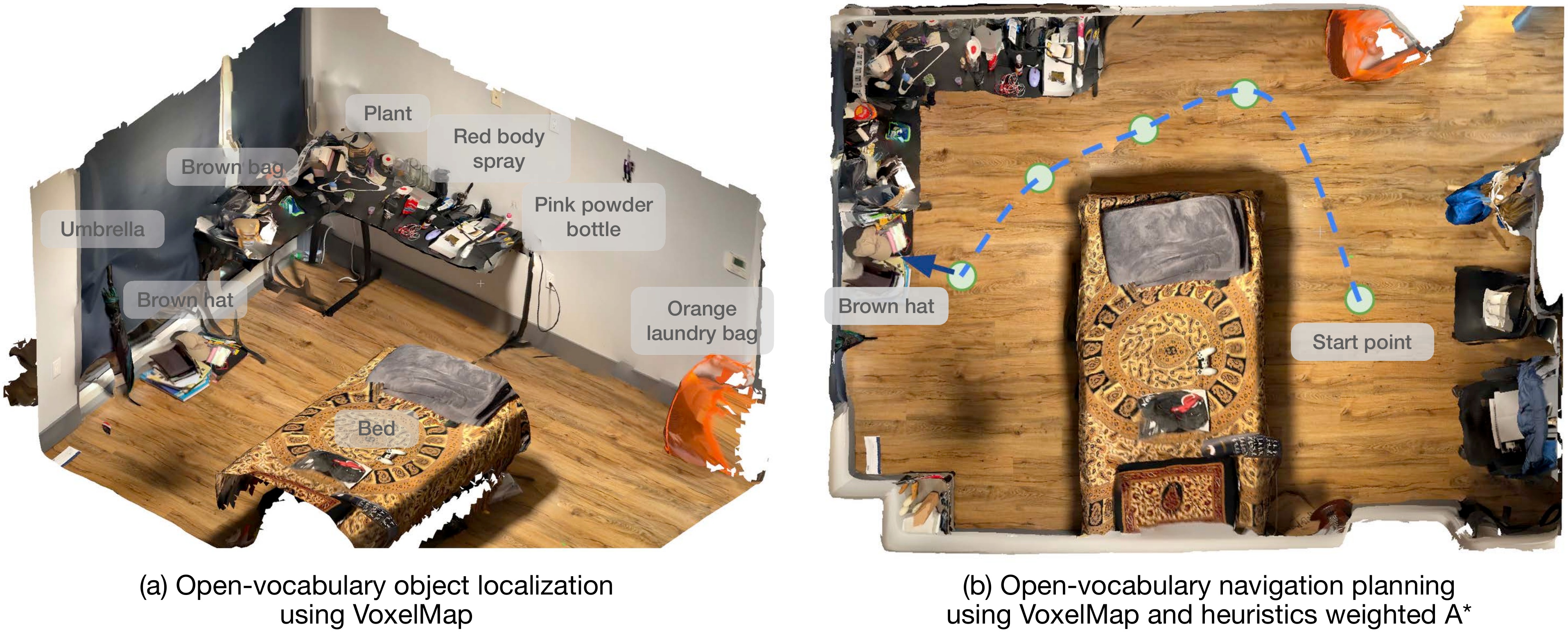}
    \caption{Open-vocabulary, open knowledge object localization and navigation in the real-world. We use the VoxelMap~\cite{homerobot} for localizing objects with natural language queries, and use an A* algorithm similar to USANet~\cite{bolte2023usa} for path planning.}
    \label{fig:objectnav}
\end{figure*}

\section{Technical Components and Method}
\label{sec:technical-comp-and-method}

Our method, on a high level, solves the problem described by the query: ``Pick up \textbf{A} (from \textbf{B}) and drop it on/in \textbf{C}'', where A is an object and B and C are places in a real-world environment such as homes. 
The system we introduce is a combination of three primary subsystems combined on a Hello Robot: Stretch.
Namely, these are the open-vocabulary object navigation module, the open-vocabulary RGB-D grasping module, and the dropping heuristic.
In this section, we describe each of these components in more details.

\subsection{Open-home, open-vocabulary object navigation}
\label{sec:objectnav}
The first component of our method is an open-home, open-vocabulary object navigation model that we use to map a home and subsequently navigate to any object of interest designated by a natural language query.

\mysection{Scanning the home}
\label{sec:collecting-robot-demo}
For open vocabulary object navigation, we follow the approach from CLIP-Fields~\cite{shafiullah2022clip} and assume a pre-mapping phase where the home is ``scanned'' manually using an iPhone. 
This manual scan simply consists of taking a video of the home using the Record3D app on the iPhone, which results in a sequence of posed RGB-D images and takes less than one minute for a new room.
Once collected, the RGB-D images, along with the camera pose and positions, are exported to our library for map-building.
To ensure our semantic memory contains both the objects of interest as well as the navigable surface and any obstacles, we capture the floor surface alongside the objects and receptacles in the environment.

\mysection{Detecting objects}
\label{sec:detecting-objects-open-vocab}
On each frame of the scan, we run an open-vocabulary object detector.
We chose OWL-ViT~\cite{minderer2022simple} over Detic~\cite{zhou2022detecting} as the object detector since we found OWL-ViT to perform better in preliminary queries.
We apply the detector on every frame, and extract each of the object bounding box, CLIP-embedding, detector confidence, and pass these information onto the object memory module.
We further refine the bounding boxes into object masks with Segment Anything (SAM)~\cite{kirillov2023segment}.
Note that, in many cases, open-vocabulary object detectors require a set of natural language object queries to be detected.
We supply a large set of such object queries, derived from the original Scannet200 labels~\cite{rozenberszki2022languagegrounded} and presented in Appendix~\ref{sec:app:scannet}, to help the detector captures most common objects in the scene.

\mysection{Object-centric semantic memory}
\label{sec:object-memory}
We use an object-centric memory similar to Clip-Fields~\cite{shafiullah2022clip} and OVMM~\cite{homerobot} that we call the VoxelMap.
VoxelMap is built by back-projecting the object masks in real-world coordinates using the depth image and the pose collected by the camera.
This process giving us a point cloud where each point has an associated CLIP semantic vector.
Then, we voxelize the point cloud to a 5 cm resolution. 
For each voxel, we calculate the detector-confidence weighted average for the CLIP embeddings that belong to that voxel.
This VoxelMap builds the base of our object memory module.
Note that the representation created this way remains static after the first scan, and cannot be adapted during the robot's operation. This inability to dynamically create a map is discussed in our limitations section (Section~\ref{sec:open-problems}).

\mysection{Querying the memory module}
\label{sec:semantic-queries}
Our semantic object memory gives us a static world representation represented as possibly non-empty voxels in the world, and a semantic vector in CLIP space associated with each voxel.
Given a language query, we first convert it to a semantic vector using the CLIP language encoder.
Then, we find the voxel where the dot product between the encoded embedding and the voxel's associated embedding is maximized.
Since each voxel is associated with a real location in the home, this lets us find the location where a queried object is most likely to be found, similar to Figure~\ref{fig:objectnav}(a).

We also implement querying for ``A on B'' by interpreting it as ``A near B''.
We do so by selecting top-10 points for query A and top-50 points for query B.
Then, we calculate the $10 \times 50$ pairwise $L_2$ distances and pick the A-point associated with the shortest (A, B) distance.
Note that during the object navigation phase we use this query only to navigate to the object approximately, and not for manipulation.
This approach gives us two advantages: our map can be as lower resolution than those in prior work~\cite{shafiullah2022clip, bolte2023usa, kerr2023lerf}, and we can deal with small movements in object's location after building the map.

\mysection{Navigating to objects in the real world}
\label{sec:gripper-tips}
Once our navigation model gives us a 3D location coordinate in the real world, we use that as a navigation target for our robot to initialize our manipulation phase.
Going and looking at an object~\cite{shafiullah2022clip, gervet2023navigating, chang2023goat} can be done while remaining at a safe distance from the object itself.
In contrast, our navigation module must place the robot at an arms length so that the robot can manipulate the target object afterwards.
Thus, our navigation method has to balance the following objectives:
\begin{enumerate}
\item The robot needs to be close enough to the object to manipulate it,
\item The robot needs some space to move its gripper, so there needs to be a small but non-negligible space between the robot and the object, and,
\item The robot needs to avoid collision during manipulation, and thus needs to keep its distance from all obstacles.
\end{enumerate}
\noindent We use three different navigation score functions, each associated with one of the above points, and evaluate them on each point of the space to find the best position to place the robot.

Let a random point be $\overrightarrow{x}$, the closest obstacle point as $\overrightarrow{x}_{obs}$, and the target object as $\overrightarrow{x_o}$. We define the following three functions $s_1, s_2, s_3$ to capture our three criterion. We define $s$ as their weighted sum. The ideal navigation point $\overrightarrow{x}^*$ is the point in space that minimizes $s(\overrightarrow{x})$, and the ideal direction is given by the vector from $\overrightarrow{x^*}$ to $\overrightarrow{x_o}$.
\begin{align*}
    s_1(\overrightarrow{x}) &= ||\overrightarrow{x} - \overrightarrow{x_o}||\\
    s_2(\overrightarrow{x}) &= 40 - \min(||\overrightarrow{x} - \overrightarrow{x_o}||, 40)\\
    s_3(\overrightarrow{x}) &= \begin{cases}
    1 / ||\overrightarrow{x} - \overrightarrow{x}_{obs}||,& \text{if } ||\overrightarrow{x} - \overrightarrow{x}_{obs}||_0\leq 30\\
    0,              & \text{otherwise}
    \end{cases}\\
    s(\overrightarrow{x}) &= s_1(\overrightarrow{x}) + 8s_2(\overrightarrow{x}) + 8s_3(\overrightarrow{x})
\end{align*}
To navigate to this target point safely from any other point in space, we follow a similar approach to~\cite{bolte2023usa, huang23avlmaps} by building an obstacle map from our captured posed RGB-D images.
We build a 2D, 10cm$\times$10cm grid of obstacles over which we navigate using the A* algorithm.
To convert our VoxelMap to an obstacle map, we first set a floor and ceiling height.
Presence of occupied voxels in between them implies the grid cell is occupied, while presence of neither ceiling nor floor voxels mean that the grid cell is unexplored.
We mark both occupied or unexplored cells as not navigable.
Around each occupied point, we mark any point within a 20 cm radius as also non-navigable to account for the robot's radius and a turn radius.
During A* search, we use the $s_3$ as a heuristic function on the node costs to navigate further away from any obstacles, which makes our generated paths similar to ideal Voronoi paths~\cite{garrido2006path} in our experiments.

\subsection{Open-vocabulary grasping in the real world}
\label{sec:grasping}

\begin{figure*}[t]
    \centering
    \includegraphics[width=\linewidth]{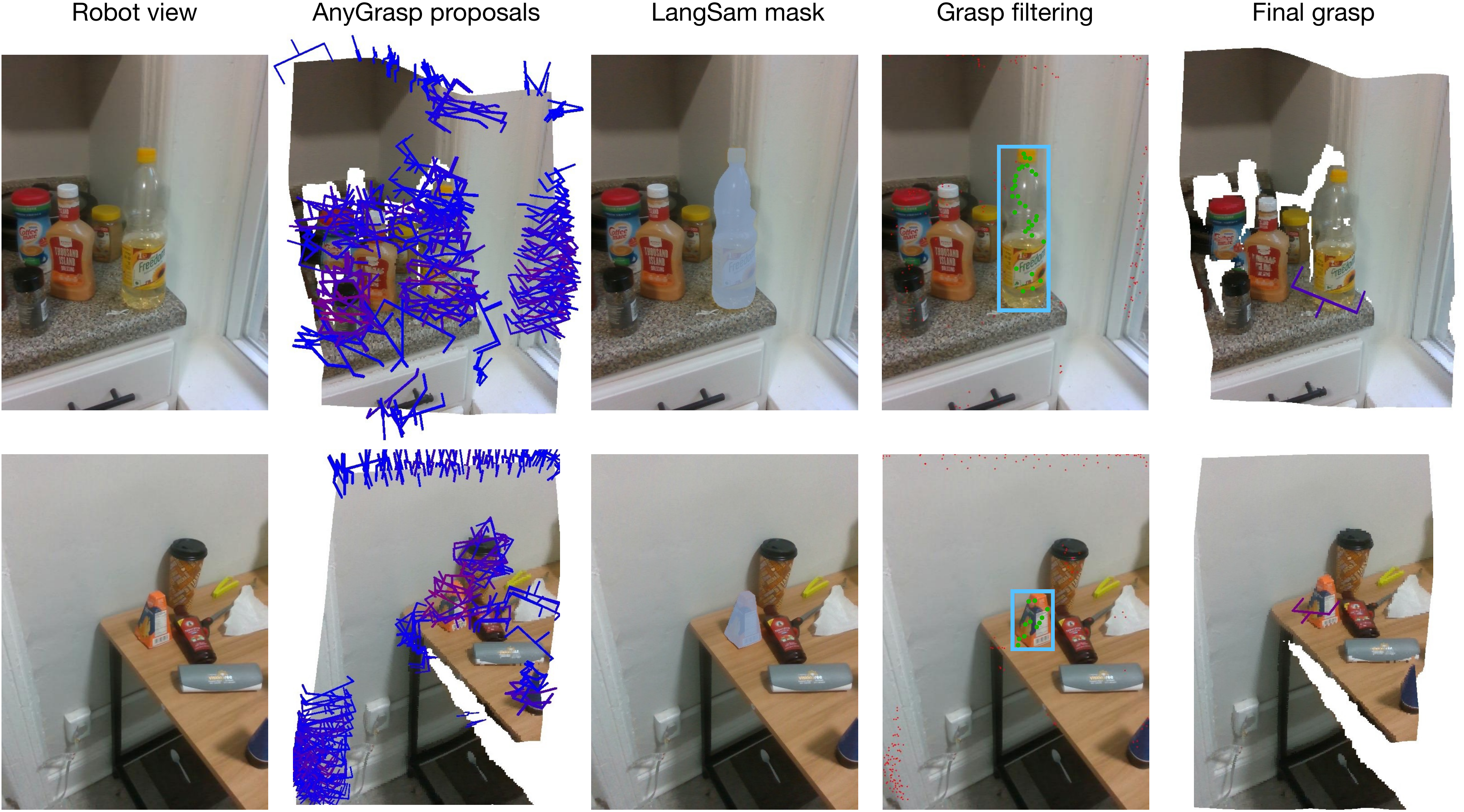}
    \caption{Open-vocabulary grasping in the real world. From left to right, we show the (a) robot POV image, (b) all suggested grasps from AnyGrasp~\cite{fang2023anygrasp}, (c) object mask given label from LangSam~\cite{langsam2023}, (d) grasp points filtered by the mask, and (e) grasp chosen for execution.}
    \label{fig:open-vocab-grasping}
\end{figure*}

Grasping or physically interacting with arbitrary objects in the real world is much more complex than open-vocabulary navigation.
We opt for using a pre-trained grasping model to generate grasp poses in the real world, and filter them with language-conditioning using a modern VLM.

\mysection{Grasp perception}
\label{sec:grasp-perception}
Once the robot reaches the object location using the navigation method outlined in Section~\ref{sec:objectnav}, we use a pre-trained grasping model, AnyGrasp~\cite{fang2023anygrasp}, to generate a grasp for the robot.
We point the robot's RGB-D head camera towards the object's 3D location, given to us by the semantic memory, and capture an RGB-D image from it (Figure~\ref{fig:open-vocab-grasping}, column 1).
We backproject and convert the depth image to a pointcloud and pass this information to the grasp generation model.
Our grasp generation model, AnyGrasp, generates all collision free grasps (Figure~\ref{fig:open-vocab-grasping} column 2) for a parallel jaw gripper in a scene given a single RGB image and a pointcloud.
AnyGrasp provides us with grasp point, width, height, depth, and a ``graspness score'', indicating uncalibrated model confidence in each grasp.

\mysection{Filtering grasps using language queries}
\label{sec:filtering-grasp}
Once we get all proposed grasps from AnyGrasp, we filter them using LangSam~\cite{langsam2023}.
LangSam~\cite{langsam2023} segments the captured image and finds the desired object mask with a language query (Figure~\ref{fig:open-vocab-grasping} column 3).
We project all the proposed grasp points onto the image and find the grasps that fall into the object mask (Figure~\ref{fig:open-vocab-grasping} column 4).
We pick the best grasp using a heuristic.
Given a grasp score $\mathcal{S}$ and the angle between the grasp normal and floor normal $\theta$, the new heuristic score is $\mathcal{S} - (\nicefrac{\theta^4}{10})$.
This heuristic balances high graspness scores with finding flat, horizontal grasps.
We prefer horizontal grasps because they are robust to small calibration errors on the robot, while vertical grasps needs better hand-eye calibration to be successful.
Robustness to hand-eye calibration errors lead to higher success as we transport the robot to different homes during our experiments.
\mysection{Grasp execution}
\label{sec:quality-control}
Once we identify the best grasp (Figure~\ref{fig:open-vocab-grasping} column 5), we use a simple pre-grasp approach~\cite{dasari2023learning} to grasp our intended object.
If $\overrightarrow{p}$ is the grasp point and $\overrightarrow{a}$ is the approach vector given by the grasping model, our robot gripper follows the following trajectory:
\begin{equation*}
    \langle \overrightarrow{p} - 0.2\overrightarrow{a},\; \overrightarrow{p} - 0.08\overrightarrow{a},\; \overrightarrow{p} - 0.04\overrightarrow{a},\; \overrightarrow{p} \rangle
\end{equation*}
Put simply, our method approaches the object from a pre-grasp position in a line with progressively smaller motions.
Moving slower as we approach the object helps the robot not knock over light objects.
Once we reach the predicted grasp point, we close the gripper in a close loop fashion to get a solid grip on the object without crushing it.
After grasping the object, we lift up the robot arm, retract it fully, and rotate the wrist to have the object tucked over the body.
This behavior maintains the robot footprint while ensuring the object is held securely by the robot and doesn't fall while navigating to the drop location.
\subsection{Dropping heuristic}
\label{sec:placing-heuristics}
After picking up an object, we find and navigatte to the drop location using the same methods described in Section~\ref{sec:objectnav}.
Unlike in HomeRobot's baseline implementation~\cite{homerobot} that assumes that the drop-off location is a flat surface, we extend our heuristic to cover concave objects such as sink, bins, boxes, and bags.
First, we segment the point cloud $P$ captured by the robot's head camera using LangSam~\cite{langsam2023} similar to Section~\ref{sec:filtering-grasp} using the drop language query.
Then, we align that segmented point cloud such that X-axis is aligned with the way the robot is facing, Y-axis is to its left and right, and the Z-axis of the point cloud is aligned with the floor normal.
Then, we normalize the point cloud so that the robot's $(x, y)$ coordinate is $(0, 0)$, and the floor plane is at $z = 0$.
We call this pointcloud $P_a$.
On the aligned, segmented point cloud, we consider the $(x, y)$ coordinates for each point, and find the median values $x_m$ and $y_m$ on each axis.
Finally, we find a drop height using $z_{\max} = 0.2 + \max \{z \mid (x, y, z) \in P_a; 0 \le x \le x_m; |y - y_m| < 0.1\}$ on the segmented, aligned pointcloud.
We add a small buffer of $0.2$ to the height to avoid collisions between the robot and the drop location.
Finally, we move the robot gripper above the drop point, and open the gripper to drop the object.
While this heuristic doesn't explicitly reason about clutter, in our experiments it performs well on average.
\subsection{Deployment in homes}
\label{sec:deployment}
Our navigation, pick, and drop primitives are combined to create our robot method that can be applied in any novel home.
For a new home environment, we ``scan'' the room in under a minute.
Then, it takes less than five minutes to process the scan into our VoxelMap.
Once that is done, the robot can be immediately placed at the base and start operating.
From arriving into a completely novel environment to start operating autonomously in it, our system takes under 10 minutes on average to complete the first pick-and-drop task.

\mysection{Transitioning between modules}
\label{sec:state-machine-model}
The transition between different modules is predefined and happens automatically once a user specifies the object to pick and where to drop it.
Since we do not implement error detection or correction, our state machine model is a simple linear chain of steps leading from navigating to object, to grasping, to navigating to goal, and to dropping the object at the goal to finish the task.

\mysection{Protocol for home experiments}
\label{sec:deployment-protocol}
To run our experiment in a novel home, we move the robot to a previously unobserved room.
We record the scene and create our VoxelMap.
Concurrently, we pick between 10-20 objects arbitrarily in each scene that can fit in the robot gripper.
These are objects found in the scene, and are not chosen ahead of time.
We come up with a language query for each chosen object using GPT-4V~\cite{openai2023gpt4} to keep the queries consistent and free of experimenter bias.
We query our navigation module to filter out all the navigation failures; i.e. objects that our semantic memory module could not locate properly.
Then, we execute pick-and-drop on remaining objects sequentially without resets between trials.

%% file: experiments.tex
\begin{figure*}[t]
    \centering
    \includegraphics[width=\linewidth]{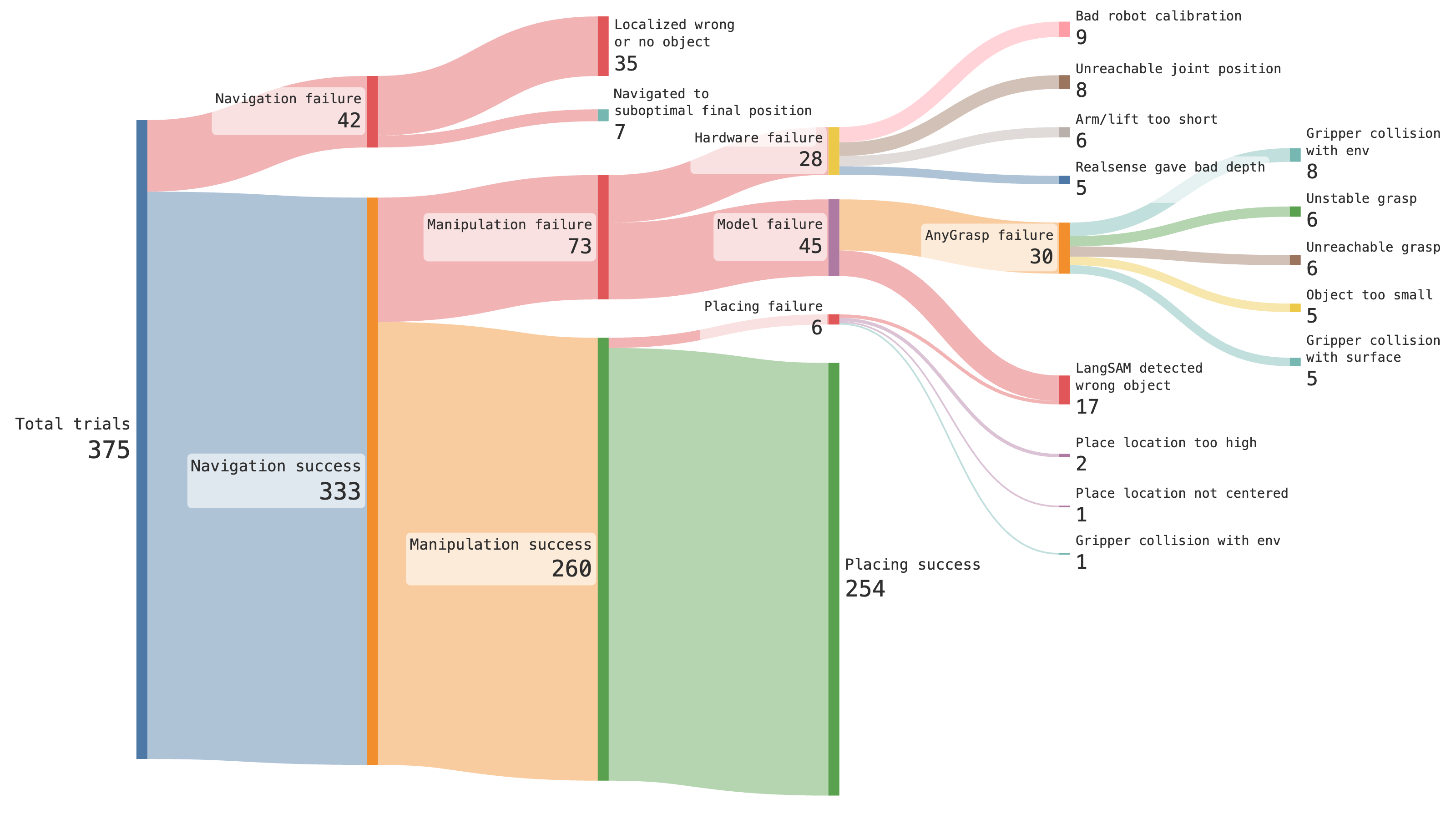}
    \caption{All the success and failure cases in our home experiments, aggregated over all three cleaning phases, and broken down by mode of failure. From left to right, we show the application of the three components of \method{}, and show a breakdown of the long-tail failure modes of each of the components.}
    \vspace{-1em}
    \label{fig:failure-mode-breakdown}
\end{figure*}

\section{Experiments}
\label{sec:experiments}
We evaluate our method in two set of experiments.
On the first set of experiments, we evaluate between multiple alternatives for each of our navigation and manipulation modules.
These experiments give us insights about which modules to use and evaluate in a home environment as a part of our method.
On the next set of experiments, we took our robots to \numhomes{} homes and ran \numhomeexperiments{} pick-and-drop experiments to empirically evaluate how our method performs in completely novel homes, and to understand the failure modes of our system.

Through these experiments, we look to answer a series of questions regarding the capabilities and limits of current Open Knowledge robotic systems, as embodied by \method{}.
Namely, we ask the following: 
\begin{enumerate}
\item How well can such a system tackle the challenge of pick and drop in arbitrary homes?
\item How well do alternate primitives for navigation and grasping compare to the recipe presented here for building an Open Knowledge robotic system?
\item How well can our current systems handle unique challenges that make homes particularly difficult, such as clutter, ambiguity, and affordance challenges? 
\item What are the failure modes of such a system and its individual components in real home environments?
\end{enumerate}

\subsection{Results of home experiments}
\label{sec:home-task-list}

Over the \numhomes{} home environment, \method{} achieved a \homesuccessrate{} success rates in completing full pick-and-drops.
Notably, this success rate is over novel objects sourced from each home with our zero-shot algorithm.
As a result, each success and failure of the robot tells us something interesting about applying open-knowledge models in robotics, which we analyze over the next sections. 
In Appendix~\ref{sec:app:home-experiments}, we provide the details of all our home experiments and results from the same.
In Appendix~\ref{sec:app:sample_objects} we show a subset of the target objects and in Appendix~\ref{sec:app:homes} we show snapshots of homes where~\method{} was deployed.
Snippets of our experiments are in Figure~\ref{fig:intro}, and full videos are presented on our project website.

\mysection{Reproduction of our system}
\label{sec:reproductions}
Beyond the home experiment results presented here, we also reproduced~\method{} in two homes in Pittsburgh, PA, and Fremont, CA.
These homes were larger and more complex: a cluttered, actively-used home kitchen environment, and a large, controlled test apartment used in prior work~\cite{homerobot,homerobotovmmchallenge2023}.
In Appendix Figure~\ref{fig:app:homes-reproductions}, we show the robot performing pick-and-drop in these two environments.
These homes were different from our initial ten experiments in a few ways.
Both were larger compared to the average NY homes, requiring more robot motion to navigate to different goals.
The PA environment (Figure~\ref{fig:app:homes-reproductions} top) notably had much more clutter.
However, given only a scan,~\method{} was able to successfully pick and drop objects like stuffed lion, plush cactus, toy drill, or green water bottle in both environments.

\begin{figure}[t]
    \centering
    \includegraphics[width=\linewidth]{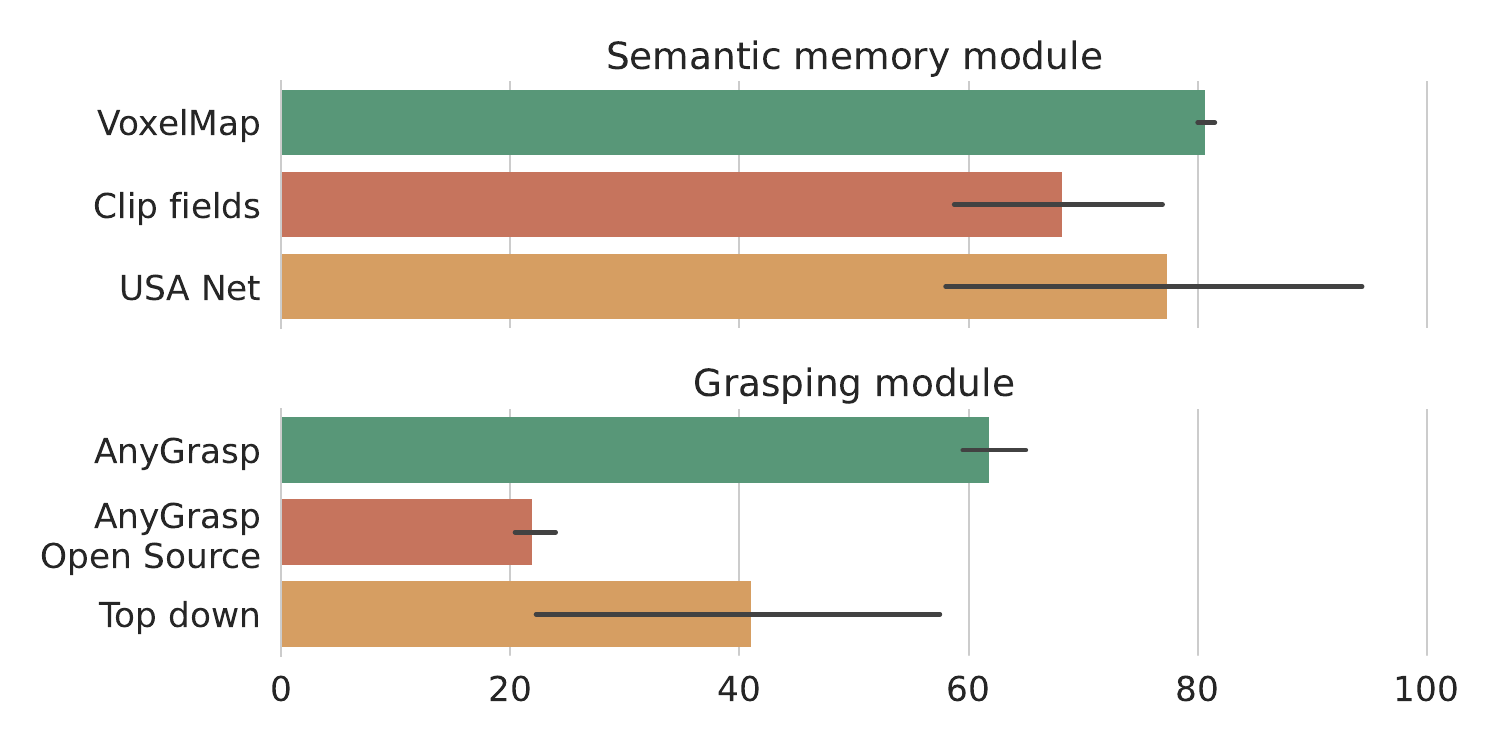}
    \caption{Ablation experiment using different semantic memory and grasping modules, with the bars showing average performance and the error bars showing standard deviation over the environments.}
    \label{fig:ablation}
    \vspace{-1em}
\end{figure}

\subsection{Ablations over system components}
\label{sec:ablations}
Apart from the navigation and manipulation strategies used in~\method{}, we also evaluated a number of alternative open vocabulary navigation and grasping modules.
We compared them by evaluating them in three different environments in our lab.
Apart from VoxelMap~\cite{homerobot}, we evaluate CLIP-Fields~\cite{shafiullah2022clip}, and USA-Net~\cite{bolte2023usa} for semantic navigation.
For grasping module, we consider AnyGrasp and its open-source baseline, Open Graspness~\cite{fang2023anygrasp}, Contact-GraspNet~\cite{sundermeyer2021contact}, and Top-Down grasp heuristic from home-robot~\cite{homerobot}.
More details about them are provided in Appendix Section~\ref{sec:app:system-components}.

In Figure~\ref{fig:ablation}, we see their comparative performance in three lab environments.
For semantic memory modules, we see that VoxelMap, used in \method{} and described in Sec.~\ref{sec:object-memory}, outperforms other semantic memory modules by a small margin.
It also has much lower variance compared to the alternatives, meaning it is more reliable.
As for grasping modules, AnyGrasp clearly outperforms other grasping methods, performing almost 50\% better in a relative scale over the next best candidate, top-down grasp.
However, the fact that a heuristic-based algorithm, top-down grasp from HomeRobot~\cite{homerobot} beats the open-source AnyGrasp baseline and Contact-GraspNet shows that building a truly general-purpose grasping model remains difficult.

\subsection{Impact of clutter, object ambiguity, and affordance}
\label{sec:by-environment-tyoe}
What makes home environments especially difficult compared to lab experiments is the presence of physical clutter, language-to-object mapping ambiguity, and hard-to-reach positions.
To gain a clear understanding of how such factors play into our experiments, we go through two ``clean-up'' processes in each environment.
During the clean-up, we pick a subset of objects that are free from ambiguity from the previous rounds, clean the clutter around objects, and generally relocated them in an accessible locations.
The two clean-up rounds at each environment gives us insights about the performance gap caused by the natural difficulties of a home-like environment.

\begin{figure}[t]
    \centering
    \includegraphics[width=\linewidth]{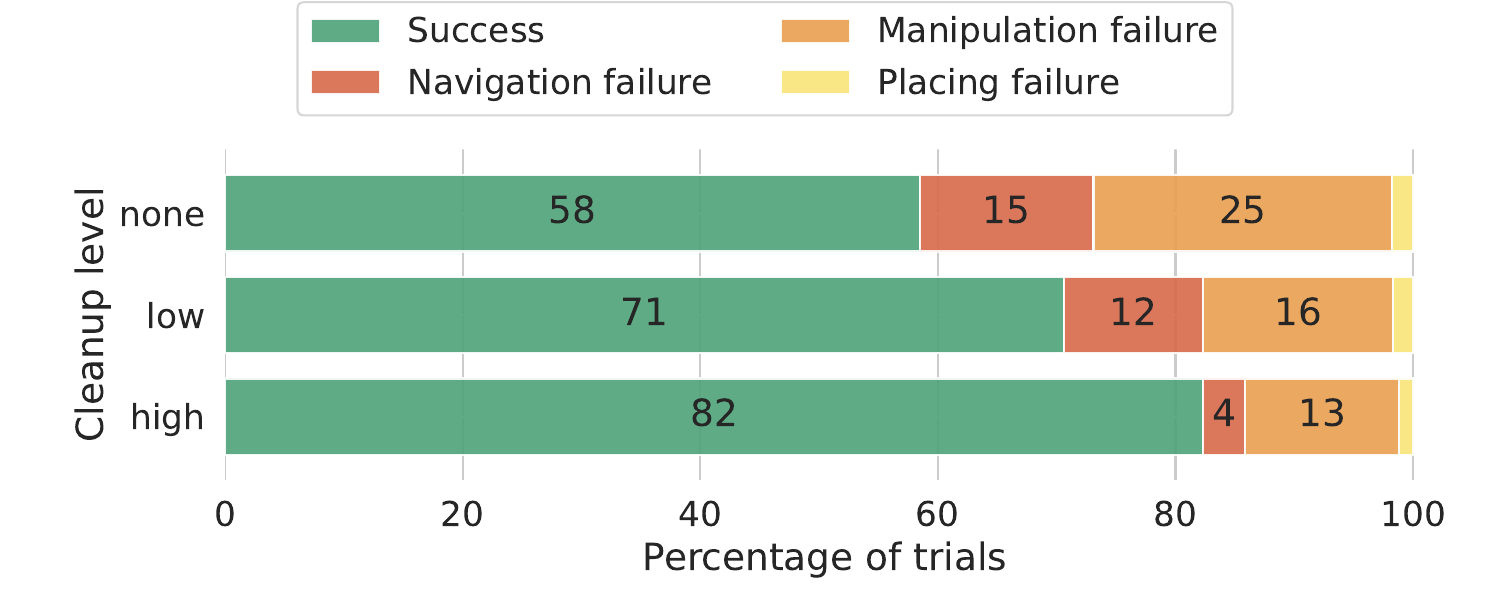}
    \caption{Failure modes of our method in novel homes, broken down by the failures of the three modules and the cleanup levels. 
    }
    \label{fig:failure-modes-by-module}
    \vspace{-1em}
\end{figure}

We show a complete analysis of the tasks listed section~\ref{sec:home-task-list} which failed in various stages in Figure~\ref{fig:failure-modes-by-module}.
As we can see from this breakdown, as we clean up the environment and remove the ambiguous objects, the navigation accuracy goes up, and the total error rate goes down from 15\% to 12\% and finally all the way down to 4\%.
Similarly, as we clean up clutters from the environment, we find that the manipulation accuracy also improves and the error rates decrease from 25\% to 16\% and finally 13\%.
Finally, since the drop-module is agnostic of the label ambiguity or manipulation difficulty arising from clutter, the failure rate of the dropping primitive stays roughly constant through the three phases of cleanup.

\subsection{Understanding the performance of \method{}}
\label{sec:understanding-performance}
While our method can show zero-shot generalization in completely new environments, we probe \method{} to better understand its failure modes.
Primarily, we elaborate on how our model performed in novel homes, what were the biggest challenges, and discuss potential solutions to them.

\label{sec:by-failure-type}
We first show a coarse-level breakdown of the failures, only considering the three high level modules of our method in Figure~\ref{fig:failure-modes-by-module}.
We see that generally, the leading cause of failure is our manipulation failure, which intuitively is the most difficult as well.
However, at a closer look, we notice a long tail of failure causes presented in figure~\ref{fig:failure-mode-breakdown}.

The three leading causes of failures are failing to retrieve the right object to navigate to from the semantic memory (9.3\%),
getting a difficult pose from the manipulation module (8.0\%),
and robot hardware difficulties (7.5\%).
In this section, we go over the analysis of the failure modes presented in Figure~\ref{fig:failure-mode-breakdown} and discuss the most frequent cases.
\label{sec:failures}
\mysection{Natural language queries for objects}
\begin{figure}[t!]
    \centering
    \includegraphics[width=\linewidth]{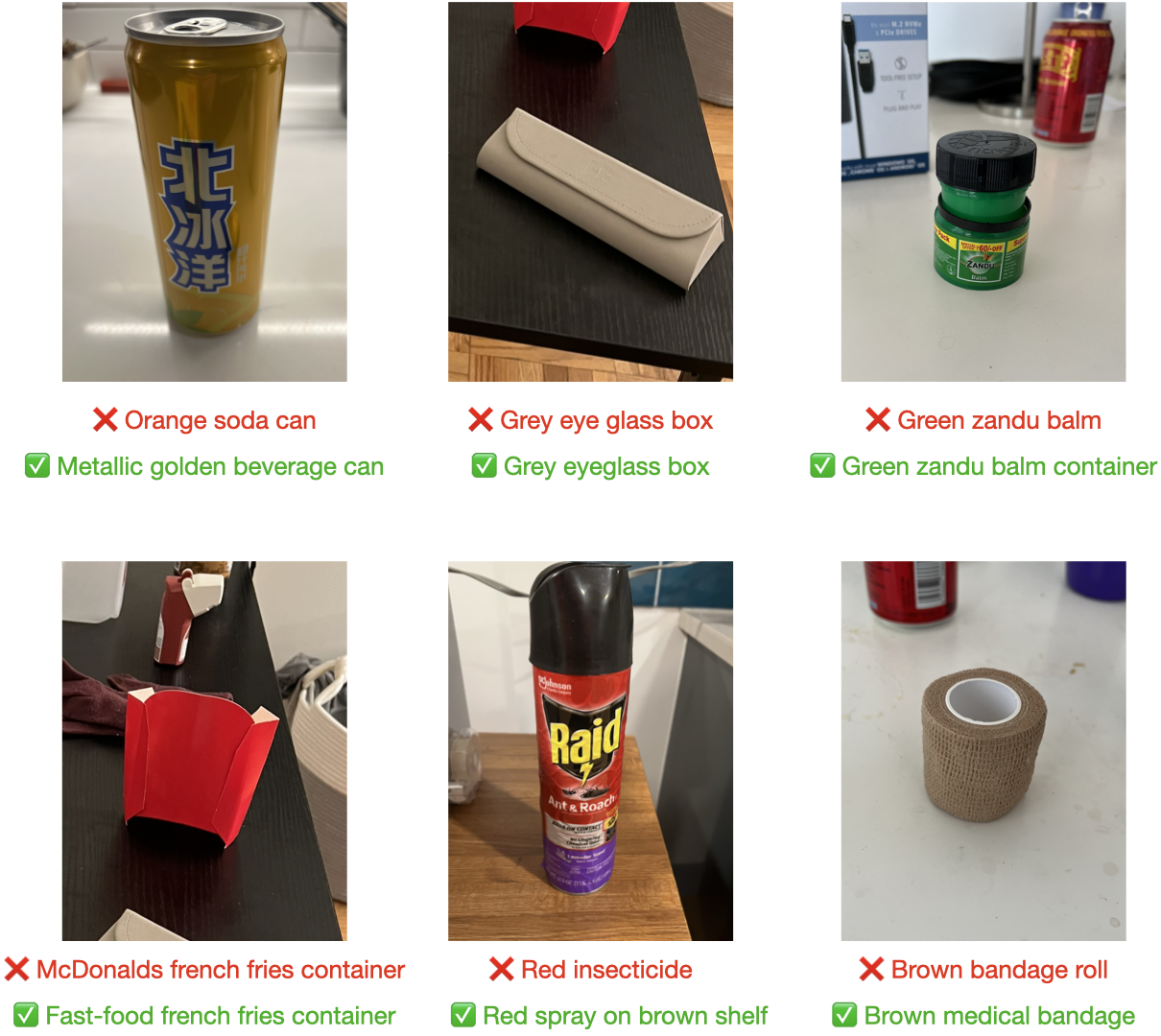}
    \caption{Samples of failed or ambiguous language queries into our semantic memory module. Since the memory module depends on pretrained large vision language model, its performance shows susceptibility to particular ``incantations'' similar to current LLMs.}
    \label{fig:failed-language-queries}
    \vspace{-1em}
\end{figure}
One of the primary reasons our \method{} can fail is when a natural language query given by the user doesn't retrieve the intended object from the semantic memory.
In Figure~\ref{fig:failed-language-queries} we show how some queries may fail while semantically very similar but slightly modified wording of the same query might succeed.

Generally, this has been the case for scenes where there are multiple visually or semantically similar objects, as shown in the figure.
There are other cases where some queries may pass while other very similar queries may fail.
An interactive system that gets confirmation from the user as it retrieves an object from memory would avoid such issues.

\begin{figure}[t!]
    \centering
    \includegraphics[width=0.95\linewidth]{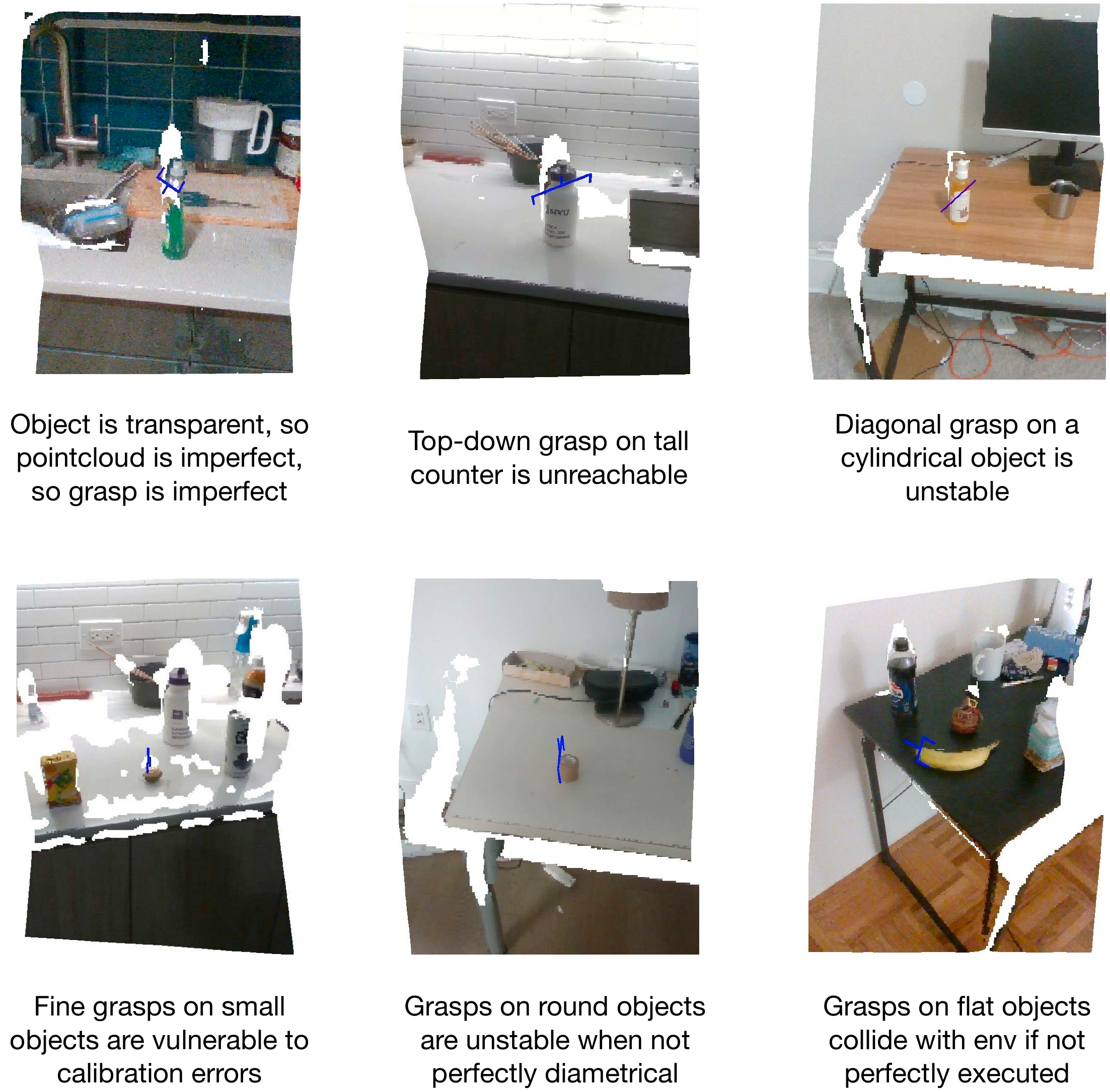}
    \caption{Samples of failures of our manipulation module. Most failures stem from using only a single RGB-D view to generate the grasp and the limiting form-factor of a large two-fingered parallel jaw gripper.}
    \label{fig:failed-grasping}
    \vspace{-1em}
\end{figure}

\mysection{Grasping module limitations}
One failure mode of our manipulation module comes from executing grasps from a pre-trained manipulation model's output based on a single RGB-D image. Moreover, this model wasn't even designed for the Hello Robot: Stretch gripper. 
As a result, sometimes the proposed grasps are unreliable or unrealistic (Figure~\ref{fig:failed-grasping}).

Sometimes, the grasp is infeasible given the robot joint limits, or is simply too far from the robot body.
Developing better grasp perception models or heuristics will let us sample better grasps for a given object.

In other cases, the model generates a good grasp pose, but as the robot is executing the grasping primitive, it collides with some minor environment obstacle.
Since we apply the same grasp trajectory in every case instead of planning the grasp trajectory, some such failures are inevitable.
Grasping models that generates a grasp trajectory as well as a pose may solve such issues.

Finally, our grasping module categorically struggles with flat objects, like chocolate bars and books, since it's difficult to grasp them off a surface with a two-fingered gripper.
\mysection{Robot hardware limitations}
\label{sec:failures:hardware-limits}
While our robot of choice, a Hello Robot: Stretch, is able to pick-and-drop a variety of objects, certain hardware limitations also dictate what our system can and cannot manipulate.
For example, the fully extended robot arm has a 1 kg payload limit, and thus our method is unable to pick objects like a full dish soap bottle.
Similarly, objects that are far from navigable floor space, i.e. in the middle of a bed, or on high places, are difficult for the robot to reach because of the reach limits of the arm.
The robot hardware or the RealSense camera can occasionally get miscalibrated over time, especially during continuous home operations. 
This miscalibration can lead to manipulation errors since that module requires hand-eye coordination in the robot.
The robot base wheels have small diameters and in some cases struggle to move smoothly between carpet and floor.

%% file: related_works.tex
\section{Related Works}
\label{sec:related-work}

\subsection{Vision-Language models for robotic navigation}
Early applications of pre-trained open-knowledge models in robotics has been in open-vocabulary navigation.
Navigating to various objects is an important task which has been looked at in a wide range of previous works~\cite{mousavian2019visual,homerobot,chang2023goat}, as well as in the context of longer pick-and-place tasks~\cite{blukis2022persistent,min2021film}.
However, these methods have generally been applied to relatively small numbers of objects~\cite{deitke2022retrospectives}.
Recently, Objaverse~\cite{deitke2023objaverse} has shown navigation to thousands of object types, for example, but much of this work has been restricted to simulated or highly controlled environments.

The early work addressing this problem builds upon representations derived from pre-trained vision language models, such as SemAbs~\cite{ha2022semantic}, CLIP-Fields~\cite{shafiullah2022clip}, VLMaps~\cite{huang2023visual}, NLMap-SayCan~\cite{chen2022nlmapsaycan}, and later, ConceptFusion~\cite{jatavallabhula2023conceptfusion} and LERF~\cite{kerr2023lerf}.
Most of these models show object localization in pre-mapped scenes, while CLIP-Fields, VLMaps, and NLMap-SayCan show integration with real robots for indoor navigation tasks.
USA-Nets~\cite{bolte2023usa} extends this task to include an affordance model, navigating with open-vocabulary queries while doing object avoidance.
ViNT~\cite{shah2023vint} proposes a foundation model for robotic navigation which can be applied to vision-language navigation problems. More recently, GOAT~\cite{chang2023goat} was proposed as a modular system for ``going to anything'' and navigating to any object in any environment given either language or image queries.
ConceptGraphs~\cite{gu2023conceptgraphs} proposed an open scene graph representation capable of handling complex queries using LLMs.
Any such open-vocabulary embodied model has the potential to improve modular systems like~\method{}.

\subsection{Pretrained robot manipulation models}
While humans can frequently  look at objects and immediately know how to grasp it, such grasping knowledge is not easily accessible to robots.
Over the years, there has been many works that has focused on creating such a general robot grasp generation model~\cite{pinto2016supersizing, gupta2018robot, Mahler2017DexNet2D, mahler2018dexnet, kalashnikov2018qt, qin2019s4g, mousavian20196dof} for arbitrary objects and potentially cluttered scenes via learning methods.
Our work focuses on more recent iterations of such methods~\cite{sundermeyer2021contact,fang2023anygrasp} that are trained on large grasping datasets~\cite{eppner2021acronym,fang2020graspnet}.
While these models only perform one task, namely grasping, they predict grasps across a large object surface and thus enable downstream complex, long-horizon manipulation tasks~\cite{goyal2022ifor,singh2023progprompt,liu2022structdiffusion}.

More recently, there is a set of general-purpose manipulation models moving beyond just grasping~\cite{shridhar2022peract, parashar2023spatial, shafiullah2022behavior,cui2022play,gervet2023act3d}. Some of these works perform general language-conditioned manipulation tasks, but are largely limited to a small set of scenes and objects.
HACMan~\cite{zhou2023learning} demonstrates a larger range of object manipulation capabilities, focused on pushing and prodding.
In the future, such models could expand the reach of our system.

\subsection{Open vocabulary robot systems}

Many recent works have worked on language-enabled tasks for complex robot systems.
Some examples include language conditioned policy learning~\cite{shridhar2021cliport, shridhar2022peract, lynch2019play, lynch2021language}, learning goal-conditioned value functions~\cite{ahn2022saycan, VoxPoser2023}, and using large language models to generate code~\cite{codeaspolicies2022, wang2023voyager, progprompt2023}.
However, a fundamental difference remains between systems which aim to operate on arbitrary objects in an open-vocab manner, and systems where one can specify one among a limited number of goals or options using language.
Consequently,  Open-Vocabulary Mobile Manipulation has been proposed as a key challenge for robotic manipulation~\cite{homerobot}.
There has previously been efforts to build such a system~\cite{ASC2023, stone2023moo}.
However, unlike such previous work, we try to build everything on an open platform and ensure our method can work without having to re-train anything for a novel home.
Recently, UniTeam~\cite{melnik2023uniteam} won the 2023 HomeRobot OVMM Challenge~\cite{homerobotovmmchallenge2023} with a modular system doing pick-and-place to arbitrary objects, with a zero-shot generalization requirement similar to ours.

In parallel, recently, there have been a number of papers doing open-vocabulary manipulation using GPT or especially GPT4~\cite{openai2023gpt4}. GPT4V can be included in robot task planning frameworks and used to execute long-horizon robot tasks, including ones from human demonstrations~\cite{wake2023gpt}. ConceptGraphs~\cite{gu2023conceptgraphs} is a good recent example, showing complex object search, planning, and pick-and-place capabilities to open-vocabulary objects. SayPlan~\cite{rana2023sayplan} also shows how these can use used together with a scene graph to handle very large, complex environments, and multi-step tasks; this work is complementary to ours, as it doesn't handle how to implement pick and place.

%% file: open_problems.tex
\section{Limitations, Open Problems and \\Requests for Research}
\label{sec:open-problems}
While our method shows significant success in completely novel home environments, it also shows many places where such methods can improve. In this section, we discuss a few of such potential improvement in the future.

\subsection{Live semantic memory and obstacle maps}
\label{sec:semantic-memory-improvements}
All the current semantic memory modules and obstacle map builders build a static representation of the world, without a good way of keeping it up-to-date as the world changes.
However, homes are dynamic environments, with many small changes over the day every day.
Future research that can build a dynamic semantic memory and obstacle map would unlock potential for continuous application of such pick-and-drop methods in a novel home out of the box.

\subsection{Grasp plans instead of proposals}
\label{sec:grasp-planners}
Currently, the grasping module proposes generic grasps without taking the robot's body and dynamics into account. 
Similarly, given a grasp pose, often the open loop grasping trajectory collides with environmental obstacles, which can be easily improved by using a module to generate grasp plans rather than grasp poses only.

\subsection{Improving interactivity between robot and user}
One of the major causes of failure in our method is in navigation: where the semantic query is ambiguous and the intended object is not retrieved from the semantic memory. In such ambiguous cases, interaction with the user would go a long way to disambiguate the query and help the robot succeed more often.

\subsection{Detecting and recovering from failure}
Currently, we observe a multiplicative error accumulation between our modules: if any of our independent components fail, the entire process fails.
As a result, even if our modules each perform independently at or above 80\% success rate, our final success rate can still be below 60\%.
However, with better error detection and retrying algorithms, we can recover from much more single-stage errors, and similarly improve our overall success rate~\cite{melnik2023uniteam}.
\subsection{Robustifying robot hardware}
While Hello Robot - Stretch~\cite{kemp2022stretch} is an affordable and portable platform on which we can implement such an open-home system for arbitrary homes, we also acknowledge that with robust hardware such methods may have vastly enhanced capacity.
Such robust hardware may enable us to reach high and low places, and pick up heavier objects.
Finally, improved robot odometry will enable us to execute much more finer grasps than is possible today.

%% file: appendix.tex
\section{Description of alternate system components}
\label{sec:app:system-components}
In this section, we provide more details about the alternate system components that we evaluated in Section~\ref{sec:ablations}.
\mysection{Alternate semantic navigation strategies}
\label{sec:ablations:navigation}
We evaluate the following semantic memory modules:
\begin{itemize}[leftmargin=12pt]
    \item \textbf{VoxelMap~\cite{homerobot}:} VoxelMap converts every detected object to a semantic vector and stores such info into an associated voxel. Occupied voxels serve as an obstacle map.
    \item \textbf{CLIP-Fields~\cite{shafiullah2022clip}:} CLIP-Fields converts a sequence of posed RGB-D images to a semantic vector field by using open-label object detectors and semantic language embedding models. The result associates each point in the space with two semantic vectors, one generated via a VLM~\cite{radford2021clip}, and another generated via a language model~\cite{reimers2019sentencebert}, which is then embedded into a neural field~\cite{mildenhall2020nerf}.
    \item \textbf{USA-Net~\cite{bolte2023usa}:} USA-Net generates multi-scale CLIP features and embeds them in a neural field that also doubles as a signed distance field. As a result, a single model can support both object retrieval and navigation.
\end{itemize}
We compare them in the same three environments with a fixed set of queries, the results of which are shown in Figure~\ref{fig:ablation}.
\mysection{Alternate grasping strategies}
\label{sec:ablations:grasping}
Similarly, we compare multiple grasping strategies to find out the best grasping strategy for our method.
\begin{itemize}[leftmargin=12pt]
    \item \textbf{AnyGrasp~\cite{fang2023anygrasp}:} AnyGrasp is a single view RGB-D based grasping model. It is trained on the GraspNet dataset which contains 1B grasp labels.
    \item \textbf{Open Graspness~\cite{fang2023anygrasp}:} Since the AnyGrasp model is free but not open source, we use an open licensed baseline trained on the same dataset.
    \item \textbf{Contact-GraspNet~\cite{sundermeyer2021contact}:} We use Contact-GraspNet as a prior work baseline, which is trained on the Acronym~\cite{eppner2021acronym} dataset. One limitation of Contact-GraspNet is that it was trained on a fixed camera view for a tabletop setting. As a result, in our application with a moving camera and arbitrary locations, it failed to give us meaningful grasps.
    \item \textbf{Top-down grasp~\cite{homerobot}:} As a heuristic based baseline, we compare with the top-down heuristic grasp provided in the HomeRobot project.
\end{itemize}

\section{Scannet200 text queries}
\label{sec:app:scannet}
To detect objects in a given home environment using OWL-ViT, we use the Scannet200 labels. The full label set is here: \texttt{['shower head', 'spray', 'inhaler', 'guitar case', 'plunger', 'range hood', 'toilet paper dispenser', 'adapter', 'soy sauce', 'pipe', 'bottle', 'door', 'scale', 'paper towel', 'paper towel roll', 'stove', 'mailbox', 'scissors', 'tape', 'bathroom stall', 'chopsticks', 'case of water bottles', 'hand sanitizer', 'laptop', 'alcohol disinfection', 'keyboard', 'coffee maker', 'light', 'toaster', 'stuffed animal', 'divider', 'clothes dryer', 'toilet seat cover dispenser', 'file cabinet', 'curtain', 'ironing board', 'fire extinguisher', 'fruit', 'object', 'blinds', 'container', 'bag', 'oven', 'body wash', 'bucket', 'cd case', 'tv', 'tray', 'bowl', 'cabinet', 'speaker', 'crate', 'projector', 'book', 'school bag', 'laundry detergent', 'mattress', 'bathtub', 'clothes', 'candle', 'basket', 'glass', 'face wash', 'notebook', 'purse', 'shower', 'power outlet', 'trash bin', 'paper bag', 'water dispenser', 'package', 'bulletin board', 'printer', 'windowsill', 'disinfecting wipes', 'bookshelf', 'recycling bin', 'headphones', 'dresser', 'mouse', 'shower gel', 'dustpan', 'cup', 'storage organizer', 'vacuum cleaner', 'fireplace', 'dish rack', 'coffee kettle', 'fire alarm', 'plants', 'rag', 'can', 'piano', 'bathroom cabinet', 'shelf', 'cushion', 'monitor', 'fan', 'tube', 'box', 'blackboard', 'ball', 'bicycle', 'guitar', 'trash can', 'hand sanitizers', 'paper towel dispenser', 'whiteboard', 'bin', 'potted plant', 'tennis', 'soap dish', 'structure', 'calendar', 'dumbbell', 'fish oil', 'paper cutter', 'ottoman', 'stool', 'hand wash', 'lamp', 'toaster oven', 'music stand', 'water bottle', 'clock', 'charger', 'picture', 'bascketball', 'sink', 'microwave', 'screwdriver', 'kitchen counter', 'rack', 'apple', 'washing machine', 'suitcase', 'ladder', 'ping pong ball', 'window', 'dishwasher', 'storage container', 'toilet paper holder', 'coat rack', 'soap dispenser', 'refrigerator', 'banana', 'counter', 'toilet paper', 'mug', 'marker pen', 'hat', 'aerosol', 'luggage', 'poster', 'bed', 'cart', 'light switch', 'backpack', 'power strip', 'baseball', 'mustard', 'bathroom vanity', 'water pitcher', 'closet', 'couch', 'beverage', 'toy', 'salt', 'plant', 'pillow', 'broom', 'pepper', 'muffins', 'multivitamin', 'towel', 'storage bin', 'nightstand', 'radiator', 'telephone', 'pillar', 'tissue box', 'vent', 'hair dryer', 'ledge', 'mirror', 'sign', 'plate', 'tripod', 'chair', 'kitchen cabinet', 'column', 'water cooler', 'plastic bag', 'umbrella', 'doorframe', 'paper', 'laundry hamper', 'food', 'jacket', 'closet door', 'computer tower', 'stairs', 'keyboard piano', 'person', 'table', 'machine', 'projector screen', 'shoe']}.

\section{Sample objects from our trials}
\label{sec:app:sample_objects}
\begin{figure*}
    \centering
    \includegraphics[width=0.9\linewidth]{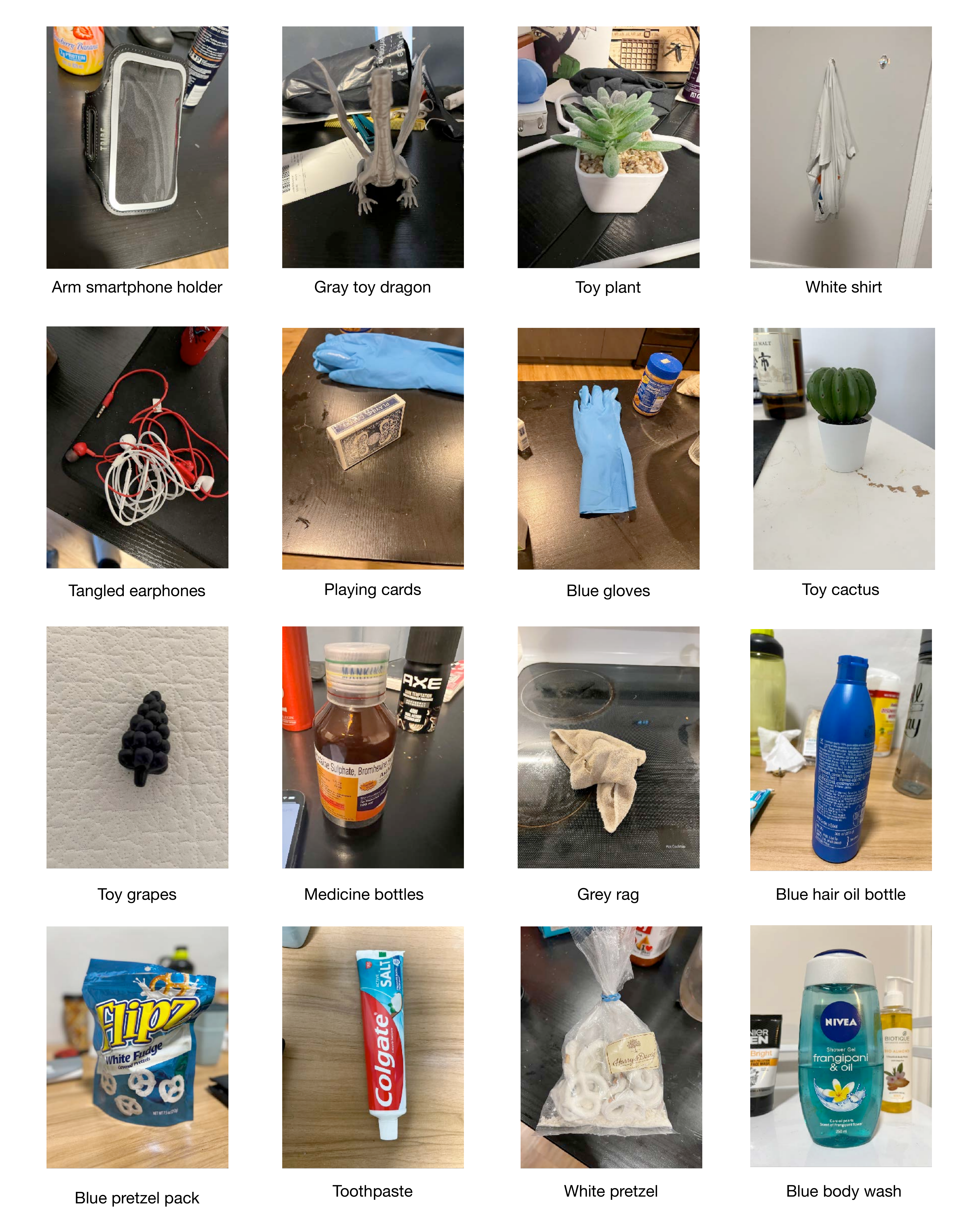}
    
    \caption{Sample objects on our home experiments, sampled from each home environment, which \method{} was able to pick and drop successfully.}
\label{fig:manip-success}
\end{figure*}

\begin{figure*}
    \centering
    \includegraphics[width=0.9\linewidth]{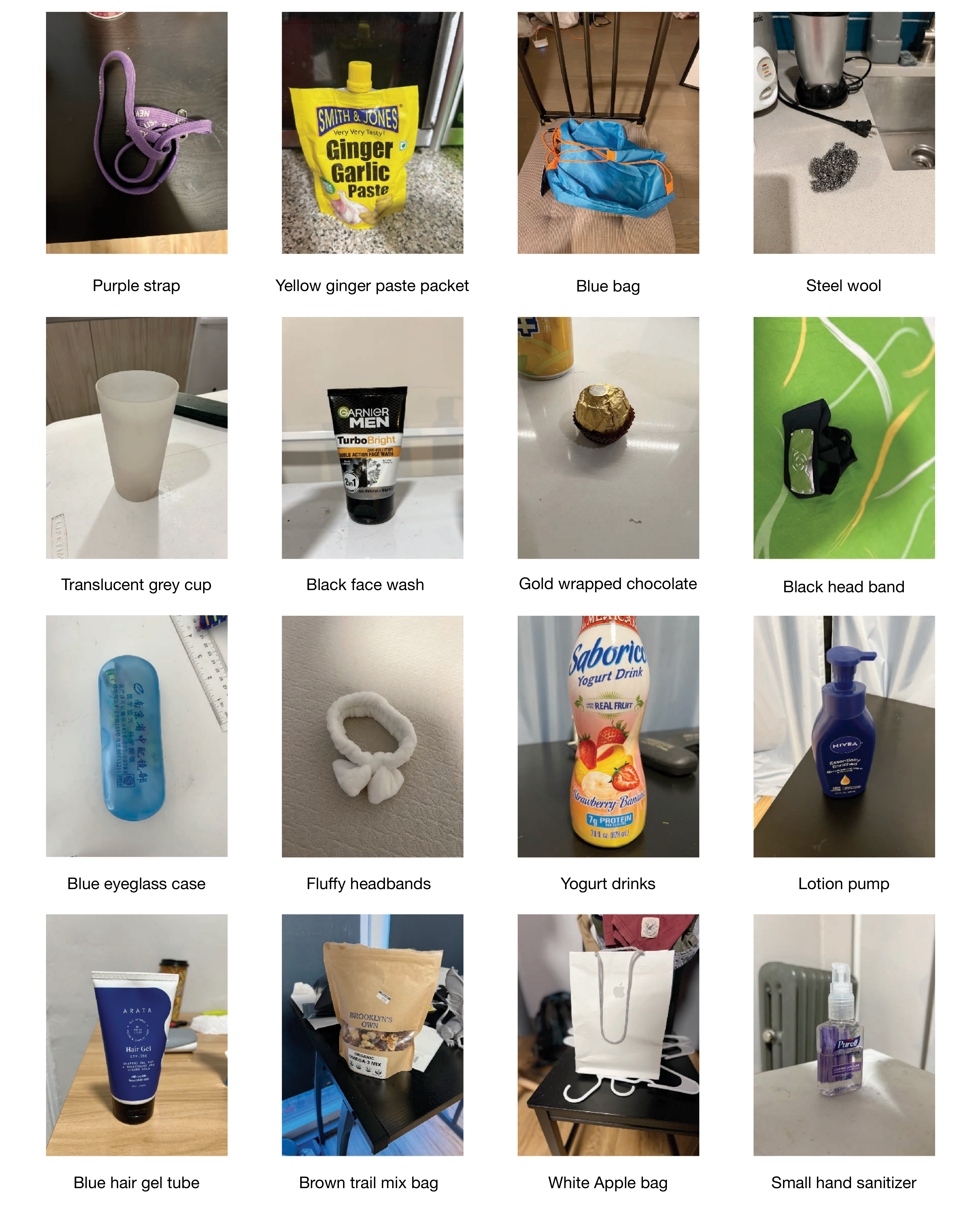}
    \caption{Sample objects on our home experiments, sampled from each home environment, which \method{} \textbf{failed} to pick up successfully.}
\label{fig:manip-failure}
\end{figure*}

During our experiments, we tried to sample objects that can plausibly be manipulated by the Hello Robot: Stretch gripper from the home environments. As a result, \method{} encountered a large variety of objects with different shapes and visual features. A subsample of such objects are presented in the Figures~\ref{fig:manip-success},~\ref{fig:manip-failure}.

\section{Sample home environments from our trials}
\label{sec:app:homes}
\begin{figure*}
    \centering
    \includegraphics[width=\linewidth]{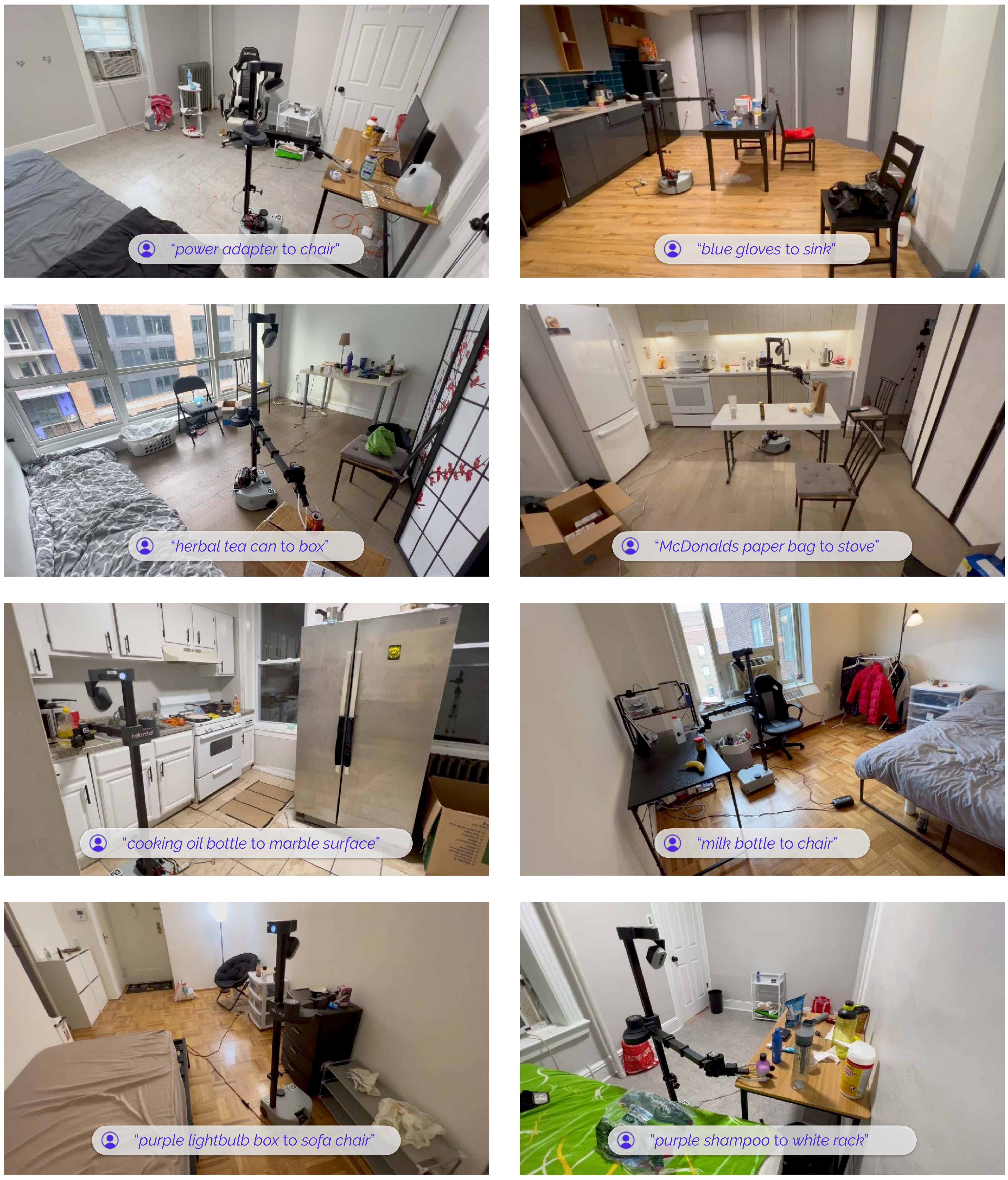}
    \caption{Eight out of the ten New York home environments in which we evaluated \method{}. In each figure caption, we show the queries that the system is being evaluated on.}
    \label{fig:app:homes-1}
\end{figure*}

\begin{figure*}
    \centering
    \includegraphics[width=0.95\linewidth]{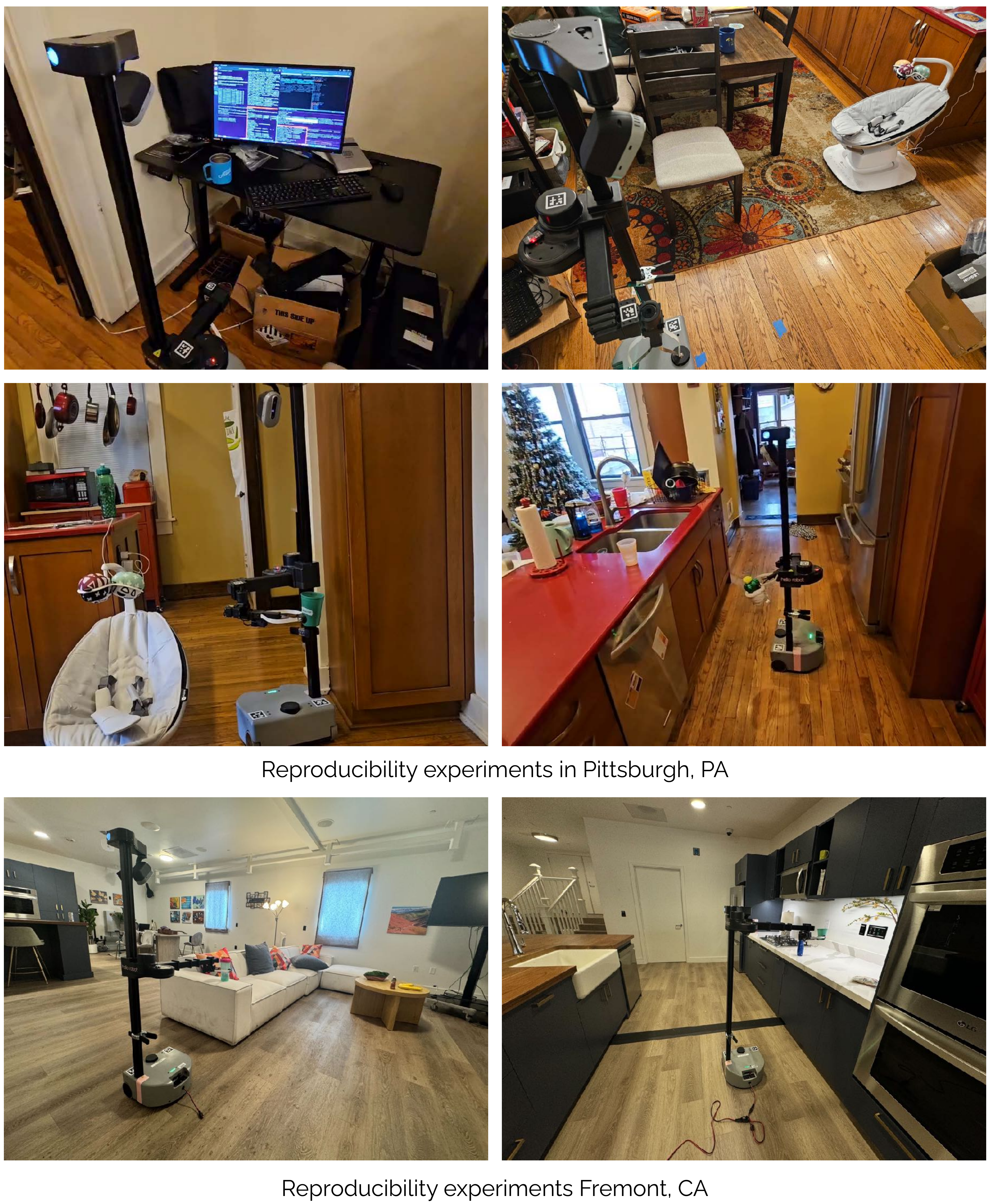}
    \caption{Home environments outside of New York where we successfully reproduced \method{}. We ensured that~\method{} can function in these homes by trying pick-and-drop on a number of objects in the homes.}
    \label{fig:app:homes-reproductions}
\end{figure*}

We show snapshots from a subset of home environments where we evaluated our method in Figures~\ref{fig:app:homes-1}. Additionally, in Figure~\ref{fig:app:homes-reproductions} we show the two home environments in Pittsburgh, PA, and Fremont, CA, where we reproduced the~\method{} system.

\section{List of home experiments}
\label{sec:app:home-experiments}
A full list of experiments in homes can be found in Table~\ref{tab:all-tasks}.

\onecolumn
\input{tables/big_table}

%% file: tables/big_table.tex
\definecolor{darkgray}{cmyk}{0.68, 0.28, 0, 0.73}
\begin{longtable}{ l c c } 
\caption{A list of all tasks in the home enviroments, along with their categories and success rates out of 10 trials.} \\
\hline
\rowcolor{darkgray}\color{white}
Pick object & \color{white}Place location & \color{white}Result                         
\\ 
\hline
\endfirsthead
\hline
\rowcolor{darkgray}\color{white}
Pick object & \color{white}Place location & \color{white}Result                         \\
\hline
\endhead
\hline
\multicolumn{3}{|r|}{\small{Continued on the next page}} \\
\hline
\endfoot
\hline
\endlastfoot
\multicolumn{3}{c}{\cellcolor[HTML]{656565}{\color[HTML]{FFFFFF} Home 1}}                                          \\ \midrule
\multicolumn{3}{c}{\cellcolor[HTML]{EFEFEF}Cleanup level: none}                                                    \\
silver cup                                      & white table                               & Success              \\
blue eye glass case                             & chair                                     & Success              \\
printed paper cup, coffee cup {[}white table{]} & \_\_\_\_                                  & Manipulation failure \\
small red and white medication                  & Chair                                     & Success              \\
power adapter                                   & Grey Bed                                  & Success              \\
wrapped paper                                   & \_\_\_\_                                         & Navigation failure   \\
\end{longtable}